\documentclass[runningheads]{llncs}
\usepackage{xcolor}
\usepackage{graphicx}
\usepackage{multirow}
\usepackage{times}
\usepackage{latexsym}
\usepackage{subcaption}
\usepackage{soul}
\usepackage{caption}
\begin{document}
%
\title{Improving Automatic Hate Speech Detection with \\ Multiword Expression Features}

%
%
\author{Nicolas Zampieri \and
Irina Illina \and
Dominique Fohr }
\authorrunning{Zampieri et al.}
%
\institute{University of Lorraine, CNRS, INRIA, Loria/ F-54000 Nancy, France
\email{\{firstname.lastname\}@loria.fr}}
\maketitle              
\begin{abstract}
The task of automatically detecting hate speech in social media is gaining more and more attention. Given the enormous volume of content posted daily, human monitoring of hate speech is unfeasible.
In this work, we propose  new word-level features for automatic hate speech detection (HSD):  multiword expressions (MWEs). MWEs are lexical units greater than a word that 
have idiomatic and compositional meanings. We propose to  integrate MWE features in a deep neural network-based HSD framework. Our baseline HSD system relies on 
Universal Sentence Encoder (USE). To incorporate MWE features, we create a three-branch deep neural network: one branch  for USE, one for MWE categories,  and one for MWE embeddings.  
We conduct experiments on two hate speech tweet corpora with different MWE categories and with two types of MWE embeddings, word2vec and BERT.
Our experiments demonstrate that the proposed HSD system with MWE features significantly outperforms the baseline system in terms of macro-F1.

\end{abstract}
\section{Introduction}


Hate speech detection (HSD) is a difficult task both for humans and machines because hateful content is more than just keyword detection. Hatred may be implied, the sentence may be grammatically incorrect and the abbreviations and slangs may be numerous \cite{Nobata2016AbusiveLD}. 
Recently, the use of machine learning methods for HSD has gained attention, as evidenced by these systems: \cite{Pamungkas2018AutomaticIO,indurthi-etal-2019-fermi}.
\cite{Lee2018} performed a comparative study
between machine learning models
and concluded that the deep learning models are more accurate.
Current HSD systems are based on natural language processing (NLP) advances and rely on deep neural networks (DNN) \cite{DBLP:journals/corr/abs-1910-12574,isaksen-gamback-2020-using}.

Finding the features that best represent the underlying hate speech phenomenon  is challenging. 
Early works on automatic HSD used different word representations, such as a bag of words, surface forms, and character n-grams with machine learning classifiers \cite{waseem-hovy-2016-hateful}. 
The combination of features, such as n-grams, linguistic and syntactic turns out to be interesting as shown by  \cite{Nobata2016AbusiveLD}. 
\cite{ychen2012} proposed lexical syntactic features, incorporating style features, structure features, and context-specific features to better predict hate speech in social media.  \cite{Chatzakou2017} investigated user features to detect bullying and aggressive behavior in tweets. 

The integration of  word embeddings, sentence embeddings, or emojis features in DNN systems allow learning semantics, contexts, and long-term dependencies.
For instance, fastText word embeddings are used in a DNN-based  HSD system \cite{Badjatiya2017DeepLF}. 
Universal Sentence Encoder  \cite{cer-etal-2018-universal} or InferSent \cite{conneau-etal-2017-supervised} allows taking into account the semantic information of the entire sentence.  
\cite{indurthi-etal-2019-fermi}  showed that sentence embeddings outperform word embeddings.  \cite{corazza-etal-2020-hybrid} proposed hybrid emoji-based Masked Language Model  to model 
the common information across different languages.  
 Convolutional Neural Network-gram based system is proposed in  \cite{rizam2020} and demonstrated   good robustness in coarse-grained and  fine-grained detection tasks.

In this paper, we focus our research on the automatic HSD in tweets using DNN. Our baseline system relies on Universal Sentence Embeddings (USE).
We propose to enrich the baseline system using word-level features, called \emph{multiword expressions} (MWEs) \cite{sag-et-al:mwe}. 
MWEs are a class of linguistic forms spanning conventional word boundaries that are both idiosyncratic and pervasive across different languages\cite{constant-etal-2017-survey}.
We believe that MWE modelling could help to reduce the ambiguity of tweets and lead to better detection of HS \cite{stankovic-etal-2020-multi}. 
To the best of our knowledge, MWE features have never been used in the framework of DNN-based automatic HSD. 
Our contribution is as follows. First, we extract different MWE categories and study their distribution in our tweet corpora. Secondly, we design a three-branch deep
neural network to integrate MWE features. To do so, the baseline system based on USE embedding is modified by adding a second branch to model different MWE categories and a third branch to
take into account the semantic meaning of MWE through word embedding. Thus, the designed system combines word-level and sentence-level features. 
Finally, we experimentally demonstrate the ability of the proposed MWE-based HSD system to better detect hate speech:
a statistically significant improvement is obtained compared to the baseline system. We experimented on two tweet corpora to show that our approach is domain-independent.

\section{Proposed methodology}
\label{sec:methodology}

In this section, we describe the proposed HSD system based on MWE features. This system is composed of  a three-branch  DNN network and combines global feature computed at the sentence level (USE embeddings) and word-level features: MWE categories and word embeddings representing  the words belonging to MWEs. 

\subsection{Universal Sentence Encoder}
\label{sec:methodology:use}
Universal sentence encoder provides sentence level embeddings. The USE model is 
trained on a variety of data sources and   demonstrated strong transfer performance on a number of NLP tasks \cite{cer-etal-2018-universal}. 
In particular, pre-trained USE showed very good performance on different  sentiment detection and semantic textual similarity tasks. 
The HSD system based on USE 
obtained the best results at the SemEval2019 campaign (shared task 5)  \cite{indurthi-etal-2019-fermi}. 
This power of USE motivated us to use it to design our  system.

\subsection{MWE features}
\label{sec:methodology:mwefeatures}
A multiword expression is a group of words that are treated as a unit \cite{sag-et-al:mwe}. For example, the two MWEs \textit{stand for} and \textit{get out} have a meaning as a group, but have another meaning if the words are taken separately. MWEs include idioms, light verb constructions, verb-particle constructions, and many compounds. 
We think that adding information about MWE categories and semantic information from MWEs might help for the HSD task. 

Automatic MWE identification and  text tagging in terms of MWEs  are difficult tasks. Different state-of-the-art deep learning methods have been studied for MWE identification, such as Convolutional Neural Network (CNN), bidirectional Long-Short Term Memory (LSTM) and transformers \cite{gharbieh-etal-2017-deep,DBLP:journals/corr/abs-1809-03056,taslimipoor-etal-2020-mtlb}. 
Over the past few years, the interest in MWEs  increased as evidenced by different shared tasks: DiMSUM \cite{schneider-etal-2016-semeval},  PARSEME  \cite{savary-etal-2017-parseme,ramisch-etal-2018-edition}. 

In our work, we focus on social media data. These textual data are very particular, may be grammatically incorrect and may contain abbreviations or spelling mistakes.
For this type of data, there are no state-of-the-art approaches for MWE identification. A specific MWE identification system is required to parse MWEs in social media corpora.  As the adaptation of an MWE identification system for a tweet corpus is a complex task and as it is not the goal of our paper, we decided to adopt a lexicon-based approach to annotate our corpora in terms of MWEs.  We extract MWEs from the STREUSLE web corpus (English online reviews corpus), annotated in MWEs \cite{schneider-smith-2015-corpus}. From this corpus, we create an MWE lexicon composed of $1855$ MWEs 
which are classified into $20$ lexical categories of MWEs.
Table \ref{mwes:examples} presents  these categories with examples.
Each tweet of our tweet corpora is lemmatized and parsed with the MWE lexicon. Our parser tags MWEs and takes into account the possible discontinuity of MWEs: we allow that one word, not belonging to the MWE, can be present between the words of the MWE. 
If, in a sentence, a word belongs to two MWEs, we tag this word with the longest MWE. We do not take into account spelling or grammatical mistakes.
We add a special category for  words not belonging to any MWE.

\subsection{HSD system proposal}
\label{sec:methodology:hsdsystem}

In this section, we describe our hate speech detection system using USE embeddings and MWE features. 
As USE is a feature at the sentence level and MWE features are at the word level, 
the architecture of our system is composed of a neural network with  three branches: two branches are dedicated to the MWE features, the last one deals with USE features.
Figure \ref{HSD-system} shows the architecture of our system.

In the first branch, we associate to each word of the tweet the number of the MWE category (one-hot encoding).
This branch is composed of 3 consecutive blocks of  CNN (Conv1D) and MaxPooling layers. 
Previous experiments with different DNN structures and the fast learning of CNN allow us to focus on this architecture. 
The second branch takes into account the semantic context of words composing MWE. If a given tweet has
one or several MWEs, we associate a word embedding 
to each word composing these MWEs. We believe that 
the semantic meaning of MWEs is important to better understand and model them. This branch uses one LSTM layer.
We propose to use two types of word embeddings: static 
where a given word has a single embedding in any context, 
or dynamic, where a given word can have different embeddings according to his long-term context.
We experiment with word2vec and BERT embeddings \cite{Mikolov:2013,devlin-etal-2019-bert}. BERT uses tokens instead of words. Therefore, we use the embedding of each token composing the words of the MWEs. 
We think that using two branches to model MWEs  allows us to take into account complementary information and provides an efficient way of combining different features for a more robust HSD system.

The last branch, USE embedding, supplies relevant semantic information at the sentence level. 
The three branches are concatenated and went through two dense layers to obtain the system output. 
The output layer has as many neurons as the number of classes to predict.
\section{Experimental setup}
\label{sec:experimentalsetup}

\subsection{Corpora}
\label{sec:experimentalsetup:corpora}
The different time frames of collection, the various sampling strategies, and the targets of abuse induce a significant shift in the data distribution and can give a performance variation on different datasets.  
We use two tweets corpora to show that our approach is domain-independent: the English corpus of SemEval2019 task 5 subTask A (called \emph{HatEval} in the following) \cite{basile-etal-2019-semeval} and Founta corpora \cite{DBLP:journals/corr/abs-1802-00393}. We study the influence of MWE features on the HatEval corpus, and we use the Founta corpus to confirm our results.
Note that these corpora contain different numbers of classes
and different percentages of hateful speech.

We evaluate our models using the official evaluation script of SemEval shared task 5 \footnote{\url{https://github.com/msang/hateval/tree/master/SemEval2019-Task5/evaluation}} in terms of macro-F1 measure. It is the average of the F1 scores of all classes.

\textbf{HatEval corpus.}
In the HatEval corpus, the annotation of a tweet is a binary value indicating if HS is occurring against women or immigrants.  The corpus contains 13k tweets. We use standard corpus partition in  training, development, and  test set with 9k, 1k, and 3k tweets respectively. Each set contains  around $42$\% of hateful tweets. The vocabulary size of the corpus is 66k words.

\begin{table}[h]
    \footnotesize
    \hskip-1.0cm
    \centering
    \captionsetup{width=0.9\linewidth}
    \caption{MWE categories with examples from STREUSLE corpus \cite{schneider-smith-2015-corpus} and the number of occurrences of MWEs. The train set of HatEvam. The column \emph{Hateful (Non-hateful)} represents MWE occurences that appear only in hateful (non-hateful) tweets. The column \emph{Both} represents MWE occurrences that appear in hateful and non-hateful tweets.}
    \begin{tabular}{l l|c|| c | c | c}
    \hline & \textbf{MWE categories} &  \textbf{Examples} & \textbf{Hateful} & \textbf{Non-hateful} & \textbf{Both}\\\hline
    \multirow{5}{*}{\rotatebox{90}{MWE5}}& Adjective &  \textit{dead on} & 9 & 8 & 255 \\
    & Adverb &  \textit{once again}  & 1 & 5 & 194 \\
   &  Discourse &  \textit{thank you} & 12 & 15 & 401 \\
   &  Nominal &  \textit{tax payer} & 25 & 36 & 189 \\
   &  Adposition phrase (idiomatic) &  \textit{on the phone} & 9 & 36 & 134\\ 
    \hline
   \multirow{5}{*}{\rotatebox{90}{VMWE5}} &  Inherently adpositional verb & \textit{stand for} & 11 & 21 & 447\\
   &  Full light verb construction & \textit{have option} & 9 & 10 & 36 \\
   &  Verbal idioms &  \textit{Give a crap} & 14 & 24 & 384 \\
   &  Full verb-particle construction &  \textit{take off} & 11 & 20 & 387 \\ 
   &  Semi verb-particle construction & \textit{walk out} & 6 & 18 & 153\\
    \hline
   &  Auxiliary &  \textit{be suppose to}  & 4 & 0 & 475 \\
  &   Coordinating conjunction & \textit{and yet} & 1 & 0 & 8\\
   &  Determiner &  \textit{a lot} & 1 & 2 & 242\\
   &  Infinitive marker &  \textit{to eat} & 0 & 0 & 12\\
   &  Adposition & \textit{apart from} & 3 & 13 & 573 \\
   &  Non-possessive pronoun &  \textit{my self} & 0 & 3 & 11 \\
    &  Subordinating conjunction & \textit{even if} & 0 & 0 & 28 \\
   &  Cause light verb construction &  \textit{give liberty} & 1 & 0 & 0\\
    \hline
    & Symbol & \textit{A+} & 0 & 0 & 0 \\
    & Interjection &  \textit{lo and behold} & 0 & 0 & 0 \\
    \hline
    \end{tabular}
    \label{mwes:examples} 
    \end{table}

 We apply the following pre-processing for each tweet: we remove mentions (words beginning by @), hashtags (words beginning by \#), and URLs. We keep the case unchanged. We use this pre-processing because the systems using this pre-processing obtained the best results at the SemEval2019 shared task 5 subtask A. 

For train and development sets, we keep only tweets that contain at least two words. Thus, we obtain  $8967$ tweets for the training set and $998$  tweets for the development set.
We split the training part into two subsets, the first one ($8003$ tweets) to train the models, and the second one ($965$ tweets) for model validation. 
In the test set, we keep all tweets after pre-processing, even empty tweets. We tag empty tweets as non-hateful. 

\textbf{Founta corpus} contains 100k tweets annotated with normal, abusive, hateful, and spam labels. Our experiments focus on HSD, so we decided to remove spams and we keep around 86k tweets. 
The vocabulary size of the corpus is 132k words.
We apply the same pre-processing as for the HatEval corpus.  We divide the Founta corpus into $3$ sets: train, development, and test with  $60$\%, $20$\%, and $20$\% respectively. As for the HatEval corpus, we use a small part of training as the validation part.  Each set contains  about $62$\%, $31$\%, and $6$\% of normal, abusive, and hateful tweets.

\subsection{System parameters}
\begin{figure}[h]
    \centering
    \includegraphics[scale=0.36]{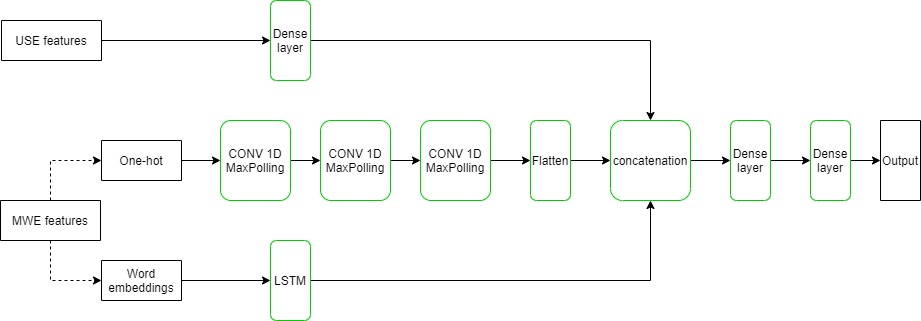}
    \caption{Proposed hate speech detection system  using USE and MWE features.}
    \label{HSD-system}
    \end{figure}

Our baseline system utilizes only USE features and 
corresponds to figure \ref{HSD-system} without
MWE branches. The system  proposed in this article  uses USE and the MWE features as presented in 
figure \ref{HSD-system} 
\footnote{https://github.com/zamp13/MWE-HSD}.  

For the USE embedding, we use the pre-trained  model provided by google\footnote{\url{https://tfhub.dev/google/universal-sentence-encoder-large/3}} (space dimension is 512) without  fine-tuning. 

We tag the MWE of each tweet using the lexicon, presented in the section \ref{sec:methodology:mwefeatures}. If an MWE is found,  we put the corresponding MWE category for all words of the MWE.
To perform fine-grained analysis, we decided to select MWE categories that have more than 50 occurrences (arbitrary value) and appear less than 97\% in hate and non-hate tweets at the same time. We obtain $10$ MWE categories: called MWE5 and VMWE5 which are respectively the first and second part of Table \ref{mwes:examples}. VMWE5 is composed of Verbal MWE categories and MWE5 with the rest of the  categories. The training part of the HatEval corpus contains  $1551$ occurrences of VMWE5 and  $1329$ occurrences of MWE5.

During our experiments, we experiment with all MWE categories presented in Table \ref{mwes:examples} (containing 19 categories: 18 categories, and a special category for words not belonging to any MWE) and with the combination of VMWE5 and MWE5 (10 MWE categories and a special category).

Concerning the MWE one-hot branch of the proposed system, we set the number of filters to 32, 16, and 8 for the 3 Conv1D layers. The kernel size of each CNN is set to 3. 

For the MWE word embedding branch, we set the LSTM layer to 192 neurons. 
 For BERT embedding, we use pre-trained uncased BERT  model from \cite{devlin-etal-2019-bert} (embedding dimension is 768).
The BERT embeddings are  extracted from the last layer of this model. BERT model is token-based, so we model each token of the words belonging to a MWE.
For word2vec embedding, we use the pre-trained embedding of  \cite{godin2019}. This model is trained on a large tweet corpus (embedding dimension is 400). 
In our systems, each  dense layer contains 256 neurons. 


For each  system configuration, we train 9 models with different random initialization. We select  the model that obtains the best result on the development set to make predictions on the test set.

\section{MWE statistics}
\label{sec:MWEstats}
\begin{figure}[h]
\centering
\includegraphics[scale=0.25]{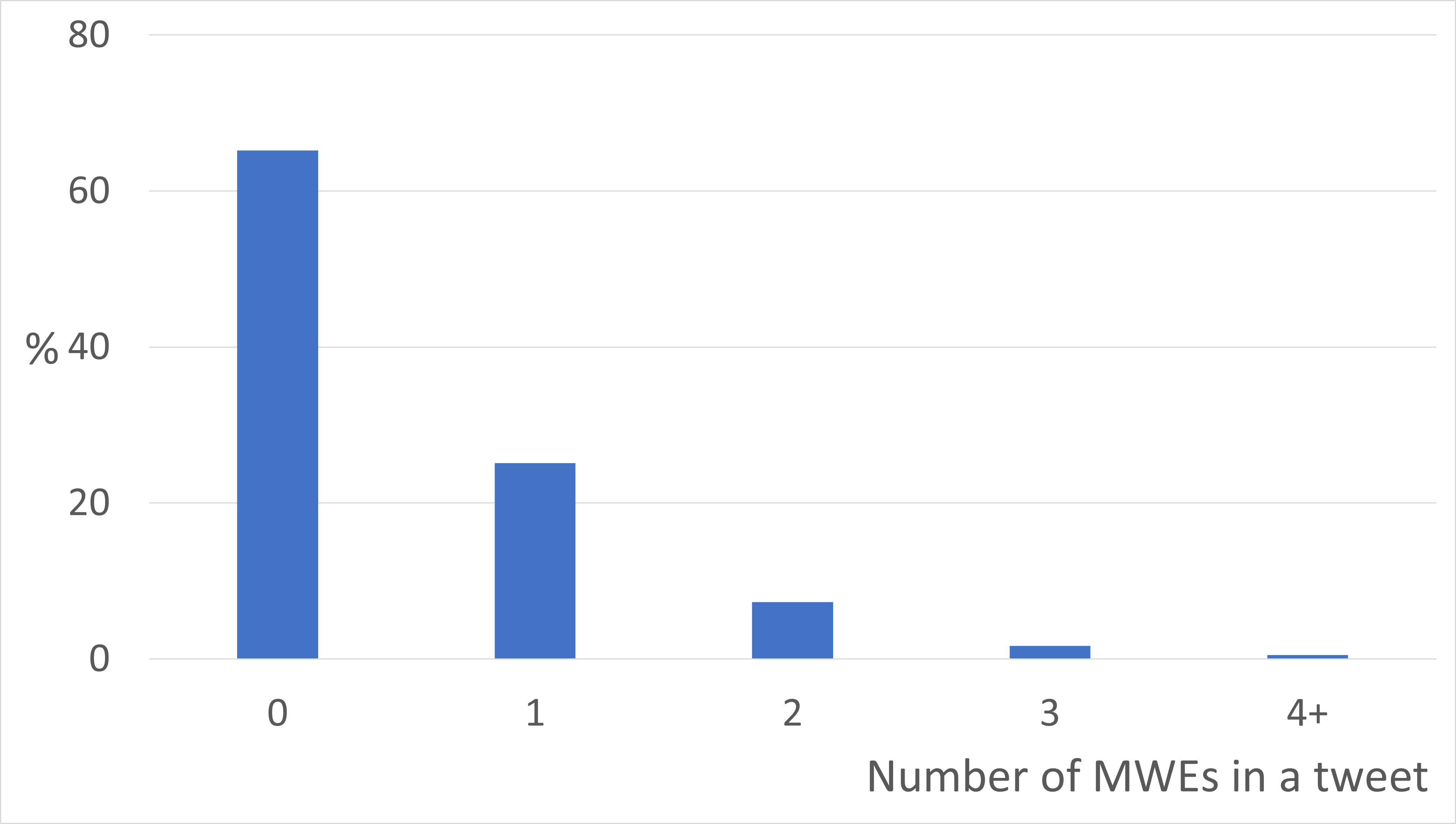}
\caption{Histogram of the occurrences of MWEs per tweet in the training set of HatEval.}
\label{MWE-per-tweet-global}
\end{figure}
We first analyze the distribution of the MWEs in our corpora.
Figure \ref{MWE-per-tweet-global} presents the percentage of occurrences of MWEs per tweet in HateEval.
We observe that about 25\% of the HatEval training tweets contain at least one MWE and enable us to
influence the HSD performance.

As a further investigation, 
we analyze MWEs appearing per MWE category and for  
hate/non-hate classes. In the training set of the HatEval corpus our parser, described in section \ref{sec:methodology:mwefeatures}, annotated $4257$ 
MWEs. Table \ref{mwes:examples} shows MWEs that appear only in hateful or non-hateful tweets or both in HatEval training part.
We observe that some MWE categories, as \emph{symbol} and \emph{interjection}, do not appear in HatEval training set.
We decided to not use these two categories in our experiments. 
Most of the categories appear in hateful and non-hateful tweets. For the majority of MWE categories, there are MWEs that occur only in hateful speech and MWEs that occur  only in non-hateful tweets.

\begin{figure}[h]
\includegraphics[width=12cm,height=4.8cm]{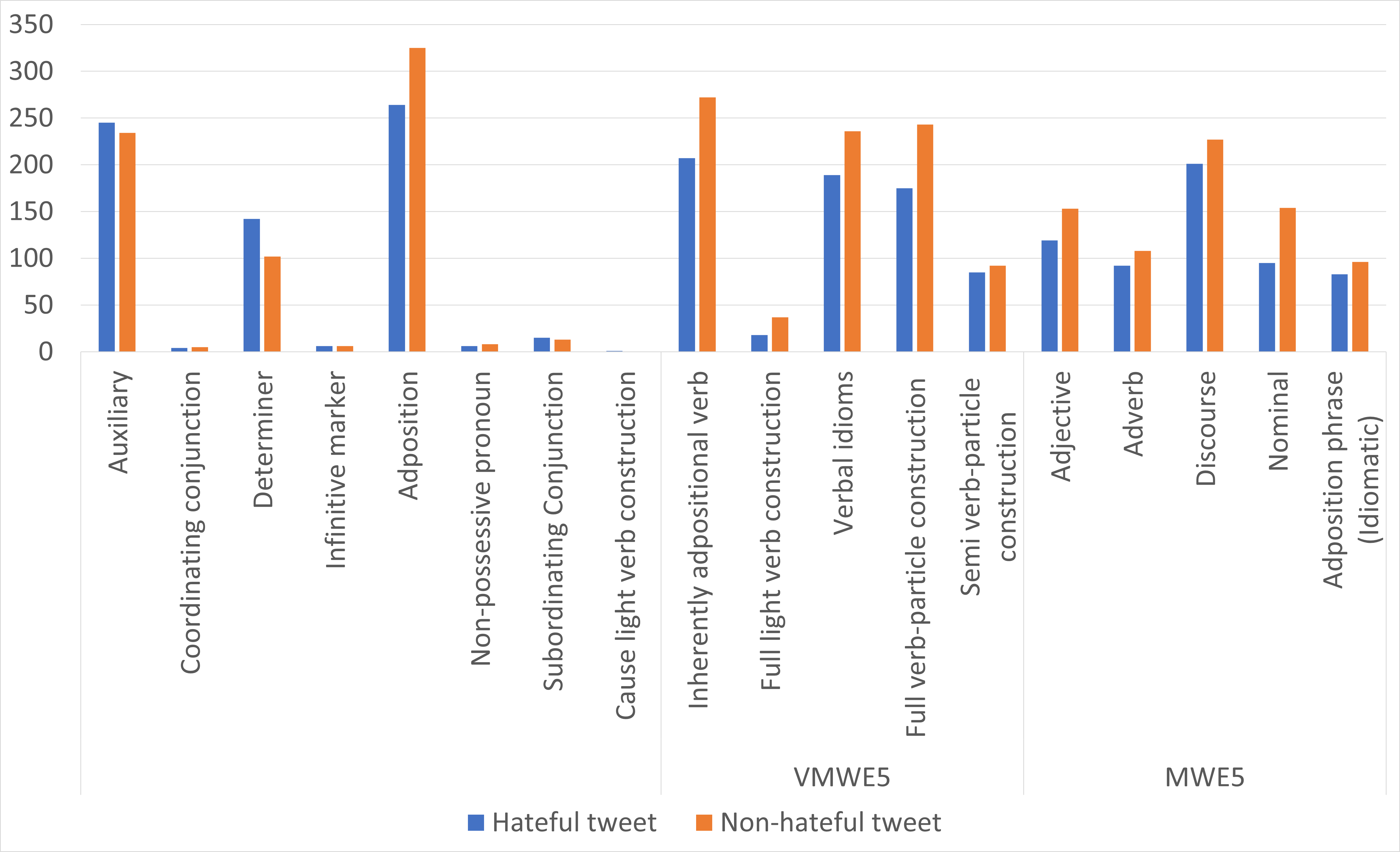}
\caption{Number of occurrences of each MWE category in the training set of HatEval. Blue (orange) bars represent occurrences in hateful (non-hateful) tweets.}
\label{MWE-stats}
\end{figure}

Figure \ref{MWE-stats} presents the statistics of each MWE category for hate and non-hate classes.
As in HatEval the classes are almost balanced  (42\% of hateful tweets, 58 \% of non-hateful tweets), there is no  bias due to imbalanced classes. 
Concerning the MWE categories, there is no categories used only in the hateful speech or only in the non-hateful speech excepted for the \emph{cause light verb construction} category, but this category is underrepresented). 
We can note that there is a difference between the use of MWEs in the hateful and the non-hateful tweets: MWEs are used more often in non-hateful speech. For some MWE categories this difference is more important, as for \emph{adposition} or \emph{full verb-particle construction}.  
In contrast, the \emph{determiner} category occurs more in hateful tweets. 
These observations reinforce our idea that MWE features can be useful for hate speech detection.

\section{Experimental results}

    

The goal of our experiments is to study the impact of MWEs on automatic hate speech detection for two different corpora: HatEval with two classes (hate and non-hate) and  Founta with three classes (hate, abusive and normal). 
We carried out experiments with the different groups of MWE categories: MWEall, including all MWE categories, and the combination of VMWE5  and MWE5.
We evaluated also different embeddings for the word embedding branch of the proposed system: word2vec and BERT. 
As described previously, we select the best-performing configuration on the development data to be applied to the test data. 

Table \ref{score-threebranches} displays the macro-F1 on HatEval and Founta test sets. Our baseline system without MWE features, called \emph{USE} in Table \ref{score-threebranches}, achieves a $65.7$\% macro-F1 score on HatEval test set. Using MWE features with word2vec or BERT embeddings,  the system proposed in this paper performs better than the baseline. For instance, on HatEval, MWEall with BERT embedding  configuration achieves the {\bf best result} with $66.8$\% of macro-F1. Regarding Founta corpus, we observe a similar result improvement: the baseline system achieves $72.2$\% and systems with MWE features obtain scores ranging from $72.4$\% to  $73.0$\% of macro-F1. It is important to note that according to a matched pair test in terms of accuracy with 5\% risk \cite{gillick-cox}, the systems using MWE features and  word2vec or BERT embeddings  \emph{significantly} outperform the baseline system on the two corpora. 
Finally,  the proposed system with MWEall and BERT embedding for HatEval outperforms the state-of-the-art system FERMI submitted at HatEval competition (SemEval task 5): $66.8$\% for our system versus $65$\% for FERMI of macro-F1 \cite{indurthi-etal-2019-fermi}. 
\begin{table}[t]
    \captionof{table}{The first part \emph{All test sets} represents F1 and macro-F1 scores (\%) on  \emph{HatEval} and \emph{Founta} test sets. The second part \emph{Tweets containing at least one MWE} represents F1 and macro-F1 scores (\%) on tweets containing at least one MWE in \emph{HatEval} and \emph{Founta} test sets.}
    \begin{tabular}{l| c c | c || c c c | c}
    
    \hline \multirow{ 3}{*}{\textbf{Features}} & \multicolumn{3}{c ||}{\textbf{HatEval}} & \multicolumn{4}{c}{\textbf{Founta}}\\
      & \multicolumn{2}{c|}{\textbf{F1}} &  \multirow{ 2}{*}{\textbf{Macro-F1}} & \multicolumn{3}{c|}{\textbf{F1}} &  \multirow{ 2}{*}{\textbf{Macro-F1}}\\
      & Hateful & Non-hate & & Norm & Abus & Hate &  \\ \hline
     \multicolumn{8}{c}{\textbf{All test sets}} \\ \hline
      USE & 64.9 & 66.4 & 65.7 & 94.2 & 87.8 & 34.6 & 72.2 \\\hline
      USE, MWEall, word2vec & 64.5 & 68.2 & 66.3 & 93.8 & 86.9 & 36.5 & 72.4\\
      USE, VMWE5, MWE5, word2vec & 66.1 & 67.0 & 66.5 & 93.9 & 87.1 & 37.2 & 72.7  \\
      \hline
      USE, MWEall, BERT & 64.2 & 69.4 & \textbf{66.8} & 94.0 & 87.1 & 37.5 & 72.9  \\
      USE, VMWE5, MWE5, BERT & 64.8 & 68.2 & 66.5 & 93.8 & 86.9 & 38.2 & \textbf{73.0} \\
    \hline
    \hline
    \multicolumn{8}{c}{\textbf{Tweets containing at least one MWE}} \\
    \hline
      USE & 67.8 & 62.3 & 65.0 & 91.1 & 94.1 & 41.6 & 75.6 \\
      USE, MWEall, word2vec & 71.7 & 61.4 & 66.6 & 91.4 & 86.9 & 44.6 & \textbf{76.5} \\
      USE, MWEall, BERT & 73.9 & 61.3 & \textbf{67.6} & 90.9 & 94 & 43.3 & 76.1\\
    \hline
    \end{tabular}
    \label{score-threebranches}
    
\end{table}
    

    

To analyze further MWE features, we experiment with different groups of MWE categories:  VMWE5, MWE5, and MWEall. 
Preliminary experiments with the two-branch system with USE and word embeddings branches only  gave a marginal improvement compared to the baseline system.  Using the three-branch neural network with only VMWE5 or MWE5 instead of MWEall seems to be interesting only for word2vec embedding. With BERT embedding it is better to use MWEall categories. Finally, the use of all MWEs could be helpful rather than the use of a subgroup of MWE categories.
Comparing word2vec and BERT embeddings, dynamic word embedding performs slightly better than the static one, however, the difference is not significant.

  \begin{table}[t]
\centering    
\caption{Confusion matrix (\%) for \emph{HatEval} test set using baseline system (USE) and proposed system (USE with MWEall and BERT embeddings).}
    \begin{tabular}{c  c | c c | c c|}
    \cline{3-6}
        &  &  \multicolumn{2}{c|}{\textbf{USE}} &  \multicolumn{2}{c|}{\textbf{USE, MWEall, BERT}} \\
         &  &  \multicolumn{2}{c|}{Predicted labels} &  \multicolumn{2}{c|}{Predicted labels} \\
         &  & Non-hateful & Hateful & Non-hateful & Hateful\\\hline
         \multirow{2}{*}{True Labels} & Non-hateful & 58.4 & 41.6 & 62.0 & 37.9 \\
         & Hateful & 24.4 & 75.6 & 27.5 & 72.4 \\
         \hline
    \end{tabular}

    \label{tab:confusionmatrix:hateval}
\end{table}
 \begin{table}[t]
     \centering
     \caption{Confusion matrix (\%) for \emph{Founta} test set using baseline system (USE) and proposed system (USE with MWEall and BERT embedding).}
    \begin{tabular}{c  c | c c c | c c c|}
    \cline{3-8}
         & &  \multicolumn{3}{c|}{\textbf{USE}} & \multicolumn{3}{c|}{\textbf{USE, MWEall, BERT}} \\
         & &  \multicolumn{3}{c|}{Predicted labels} & \multicolumn{3}{c|}{Predicted labels} \\
         &  & Non-hate & Abusive & Hateful & Non-hate & Abusive & Hateful \\\hline
         \multirow{3}{3em}{True labels} & Non-hateful & 95.4  & 4.1 & 0.4 & 95.2 & 4.2 & 0.4\\
         & Abusive & 7.5 & 90.1 & 2.3 & 8.7 & 88.5 & 2.6 \\
         & Hateful & 35.4 & 39.8 & 24.7 & 34.8 & 37.6 & 27.5\\
         \hline
    \end{tabular}
    \label{tab:confusionmatrix:founta}
\end{table}  
Table \ref{tab:confusionmatrix:hateval}  compares the confusion matrices of two systems: the baseline system and the proposed one with MWEall and BERT embeddings. On the HatEval test corpus, the proposed system classifies better non-hateful tweets than the baseline system (62.0\% versus 58.4\%). 
On the other hand, the proposed system classifies a little less well hate speech (72.4\% versus 75.6 \%). 
On Founta test set (see Table \ref{tab:confusionmatrix:founta}),  
the conclusions are different, the proposed system  classifies better hateful tweets than baseline system (27.5\% versus 24.7\%), and a little less well normal and abusive speech: 95.2\% versus 95.4\% for normal speech and 88.5\% versus 90.1\% for abusive speech. This difference in detection results between HatEval and Founta is confirmed by the F1 score per class in the first part of table  \ref{score-threebranches}.
We think that the balance between the classes plays an important role here: in the case of HatEval corpus,  the classes are balanced, in the case of Founta, the classes are unbalanced. 

To perform a deeper analysis, we focus our observations on only the tweets from the test sets containing at least one MWE: 758 tweets from  the HatEval test set and 3508 tweets from the Founta test set. Indeed, according to section \ref{sec:MWEstats}, there is about 25\% of tweets containing MWEs.
The second part of table \ref{score-threebranches} shows that  the results are consistent with those observed previously in this section, and the obtained improvement is more important. 

\section{Conclusion}
In this work, we explored a new way to design a HSD system for short texts, like tweets. We proposed to add new features to our DNN-based detection system: mutliword expression features. 
 We integrated MWE features in a USE-based neural network thanks to  a neural network of three branches. This network allows to take into account sentence-level features (USE embedding) and word-level features (MWE categories and the embeddings of  the words belonging to the MWEs). 
 The  results were validated on two tweet corpora: HatEval and Founta. 
 The models we proposed yielded significant improvements in macro-F1 over the baseline system (USE system).
Furthermore, on HatEval corpus, the proposed system with MWEall categories and BERT embedding significantly outperformed the state-of-the-art system FERMI ranked first at the SemEval2019 shared task 5. 
These results showed that MWE features allow to enrich our baseline system. 
The proposed approach can be adapted to other NLP tasks, like sentiment analysis or automatic translation. 


\bibliographystyle{splncs03}
\bibliography{arxiv}

\begin{thebibliography}{10}
\providecommand{\url}[1]{\texttt{#1}}
\providecommand{\urlprefix}{URL }
\providecommand{\doi}[1]{https://doi.org/#1}

\bibitem{Badjatiya2017DeepLF}
Badjatiya, P., Gupta, S., Gupta, M., Varma, V.: Deep learning for hate speech
  detection in tweets. Proceedings of the 26th International Conference on
  World Wide Web Companion  (2017)

\bibitem{basile-etal-2019-semeval}
Basile, V., Bosco, C., Fersini, E., Nozza, D., Patti, V., Rangel~Pardo, F.M.,
  Rosso, P., Sanguinetti, M.: {S}em{E}val-2019 task 5: Multilingual detection
  of hate speech against immigrants and women in twitter. In: Proceedings of
  the 13th International Workshop on Semantic Evaluation. pp. 54--63. ACL,
  Minneapolis, Minnesota, USA (Jun 2019). \doi{10.18653/v1/S19-2007},
  \url{https://www.aclweb.org/anthology/S19-2007}

\bibitem{cer-etal-2018-universal}
Cer, D., Yang, Y., Kong, S.y., Hua, N., Limtiaco, N., St.~John, R., Constant,
  N., Guajardo-Cespedes, M., Yuan, S., Tar, C., Strope, B., Kurzweil, R.:
  Universal sentence encoder for {E}nglish. In: Proceedings of the 2018
  Conference on Empirical Methods in Natural Language Processing: System
  Demonstrations. pp. 169--174. ACL, Brussels, Belgium (Nov 2018).
  \doi{10.18653/v1/D18-2029}, \url{https://www.aclweb.org/anthology/D18-2029}

\bibitem{Chatzakou2017}
Chatzakou, D., Kourtellis, N., Blackburn, J., Cristofaro, E.D., Stringhini, G.,
  Vakali, A.: Mean birds:detecting aggression and bullying on twitter. In:
  Proceedings of the 2017 ACM on Web Science Conference. p. 13–22 (2017)

\bibitem{ychen2012}
Chen, Y., Zhu, S., Zhou, Y., Xu, H.: Detecting offensive language in social
  media to protect adolescent online safety. In: Proceedings of the 2012
  ASE/IEEE International Conference on Social Computing. pp. 71--–80.
  Washington, USA (2012)

\bibitem{conneau-etal-2017-supervised}
Conneau, A., Kiela, D., Schwenk, H., Barrault, L., Bordes, A.: Supervised
  learning of universal sentence representations from natural language
  inference data. In: Proceedings of the 2017 Conference on Empirical Methods
  in Natural Language Processing. pp. 670--680. ACL, Copenhagen, Denmark (Sep
  2017). \doi{10.18653/v1/D17-1070},
  \url{https://www.aclweb.org/anthology/D17-1070}

\bibitem{constant-etal-2017-survey}
Constant, M., Eryi{\v{g}}it, G., Monti, J., Plas, L.v.d., Ramisch, C., Rosner,
  M., Todirascu, A.: Multiword expression processing: A {S}urvey. Computational
  Linguistics  \textbf{43}(4),  837--892 (Dec 2017),
  \url{https://www.aclweb.org/anthology/J17-4005}

\bibitem{corazza-etal-2020-hybrid}
Corazza, M., Menini, S., Cabrio, E., Tonelli, S., Villata, S.: Hybrid
  emoji-based masked language models for zero-shot abusive language detection.
  In: Findings of the Association for Computational Linguistics: EMNLP 2020.
  pp. 943--949. ACL, Online (Nov 2020),
  \url{https://www.aclweb.org/anthology/2020.findings-emnlp.84}

\bibitem{devlin-etal-2019-bert}
Devlin, J., Chang, M.W., Lee, K., Toutanova, K.: {BERT}: Pre-training of deep
  bidirectional transformers for language understanding. In: Proceedings of the
  2019 Conference of the North {A}merican Chapter of the Association for
  Computational Linguistics: Human Language Technologies, Volume 1 (Long and
  Short Papers). pp. 4171--4186. ACL, Minneapolis, Minnesota (Jun 2019).
  \doi{10.18653/v1/N19-1423}, \url{https://www.aclweb.org/anthology/N19-1423}

\bibitem{DBLP:journals/corr/abs-1802-00393}
Founta, A., Djouvas, C., Chatzakou, D., Leontiadis, I., Blackburn, J.,
  Stringhini, G., Vakali, A., Sirivianos, M., Kourtellis, N.: Large scale
  crowdsourcing and characterization of twitter abusive behavior. CoRR
  \textbf{abs/1802.00393} (2018), \url{http://arxiv.org/abs/1802.00393}

\bibitem{gharbieh-etal-2017-deep}
Gharbieh, W., Bhavsar, V., Cook, P.: Deep learning models for multiword
  expression identification. In: Proceedings of the 6th Joint Conference on
  Lexical and Computational Semantics (*{SEM} 2017). pp. 54--64. ACL,
  Vancouver, Canada (Aug 2017). \doi{10.18653/v1/S17-1006},
  \url{https://www.aclweb.org/anthology/S17-1006}

\bibitem{gillick-cox}
{Gillick}, L., {Cox}, S.J.: Some statistical issues in the comparison of speech
  recognition algorithms. In: International Conference on Acoustics, Speech,
  and Signal Processing,. pp. 532--535 vol.1. Glasgow, UK (1989)

\bibitem{godin2019}
Godin, F.: Improving and Interpreting Neural Networks for Word-Level Prediction
  Tasks in Natural Language Processing. Ph.D. thesis, Ghent University, Belgium
  (2019)

\bibitem{indurthi-etal-2019-fermi}
Indurthi, V., Syed, B., Shrivastava, M., Chakravartula, N., Gupta, M., Varma,
  V.: {FERMI} at {S}em{E}val-2019 task 5: Using sentence embeddings to identify
  hate speech against immigrants and women in twitter. In: Proceedings of the
  13th International Workshop on Semantic Evaluation. pp. 70--74. ACL,
  Minneapolis, Minnesota, USA (Jun 2019). \doi{10.18653/v1/S19-2009},
  \url{https://www.aclweb.org/anthology/S19-2009}

\bibitem{isaksen-gamback-2020-using}
Isaksen, V., Gamb{\"a}ck, B.: Using transfer-based language models to detect
  hateful and offensive language online. In: Proceedings of the Fourth Workshop
  on Online Abuse and Harms. pp. 16--27. ACL, Online (Nov 2020),
  \url{https://www.aclweb.org/anthology/2020.alw-1.3}

\bibitem{Lee2018}
Lee, Y., Yoon, S., Jung, K.: Comparative studies of detecting abusive language
  on twitter. In: Proceedings of the 2nd Workshop on Abusive Language Online
  (ALW2). p. 101–106 (2018)

\bibitem{Mikolov:2013}
Mikolov, T., Chen, K., Corrado, G., Dean, J.: Efficient estimation of word
  representations in vector space. In: ICLR Workshop Papers (2013),
  \url{http://arxiv.org/abs/1301.3781}

\bibitem{DBLP:journals/corr/abs-1910-12574}
Mozafari, M., Farahbakhsh, R., Crespi, N.: A bert-based transfer learning
  approach for hate speech detection in online social media. CoRR
  \textbf{abs/1910.12574} (2019), \url{http://arxiv.org/abs/1910.12574}

\bibitem{Nobata2016AbusiveLD}
Nobata, C., Tetreault, J., Thomas, A., Mehdad, Y., Chang, Y.: Abusive language
  detection in online user content. In: Proceedings of the 25th International
  Conference on World Wide Web. p. 145–153. Montréal, Canada (2016)

\bibitem{Pamungkas2018AutomaticIO}
Pamungkas, E.W., Cignarella, A.T., Basile, V., Patti, V.: Automatic
  identification of misogyny in english and italian tweets at evalita 2018 with
  a multilingual hate lexicon. In: EVALITA@CLiC-it (2018)

\bibitem{ramisch-etal-2018-edition}
Ramisch, C., Cordeiro, S.R., Savary, A., Vincze, V., Barbu~Mititelu, V.,
  Bhatia, A., Buljan, M., Candito, M., Gantar, P., Giouli, V., G{\"u}ng{\"o}r,
  T., Hawwari, A., I{\~n}urrieta, U., Kovalevskait{\.e}, J., Krek, S., Lichte,
  T., Liebeskind, C., Monti, J., Parra~Escart{\'\i}n, C., QasemiZadeh, B.,
  Ramisch, R., Schneider, N., Stoyanova, I., Vaidya, A., Walsh, A.: Edition 1.1
  of the {PARSEME} shared task on automatic identification of verbal multiword
  expressions. In: Proceedings of the Joint Workshop on Linguistic Annotation,
  Multiword Expressions and Constructions ({LAW}-{MWE}-{C}x{G}-2018). pp.
  222--240. ACL, Santa Fe, New Mexico, USA (Aug 2018),
  \url{https://www.aclweb.org/anthology/W18-4925}

\bibitem{rizam2020}
Rizwan, H., Shakeel, M.H., Karim, A.: Hate-speech and offensive language
  detection in roman urdu. In: Proceedings of the 2020 Conference on Empirical
  Methods in Natural Language Processing EMNLP. pp. 2512--2522 (2020)

\bibitem{sag-et-al:mwe}
Sag, I., Baldwin, T., Bond, F., Copestake, A., Flickinger, D.: Multiword
  expressions: A pain in the neck for nlp. In: Proceedings of CICLING-2002. pp.
  1--15. Mexico City, Mexico (02 2002)

\bibitem{savary-etal-2017-parseme}
Savary, A., Ramisch, C., Cordeiro, S., Sangati, F., Vincze, V., QasemiZadeh,
  B., Candito, M., Cap, F., Giouli, V., Stoyanova, I., Doucet, A.: The
  {PARSEME} shared task on automatic identification of verbal multiword
  expressions. In: Proceedings of the 13th Workshop on Multiword Expressions
  ({MWE} 2017). pp. 31--47. ACL, Valencia, Spain (Apr 2017).
  \doi{10.18653/v1/W17-1704}, \url{https://www.aclweb.org/anthology/W17-1704}

\bibitem{schneider-etal-2016-semeval}
Schneider, N., Hovy, D., Johannsen, A., Carpuat, M.: {S}em{E}val-2016 task 10:
  Detecting minimal semantic units and their meanings ({D}i{MSUM}). In:
  Proceedings of the 10th International Workshop on Semantic Evaluation
  ({S}em{E}val-2016). pp. 546--559. ACL, San Diego, California (Jun 2016).
  \doi{10.18653/v1/S16-1084}, \url{https://www.aclweb.org/anthology/S16-1084}

\bibitem{schneider-smith-2015-corpus}
Schneider, N., Smith, N.A.: A corpus and model integrating multiword
  expressions and supersenses. In: Proceedings of the 2015 Conference of the
  North {A}merican Chapter of the Association for Computational Linguistics:
  Human Language Technologies. pp. 1537--1547. ACL, Denver, Colorado
  (May{--}Jun 2015). \doi{10.3115/v1/N15-1177},
  \url{https://www.aclweb.org/anthology/N15-1177}

\bibitem{stankovic-etal-2020-multi}
Stankovi{\'c}, R., Mitrovi{\'c}, J., Joki{\'c}, D., Krstev, C.: Multi-word
  expressions for abusive speech detection in {S}erbian. In: Proceedings of the
  Joint Workshop on Multiword Expressions and Electronic Lexicons. pp. 74--84.
  ACL, online (Dec 2020), \url{https://www.aclweb.org/anthology/2020.mwe-1.10}

\bibitem{taslimipoor-etal-2020-mtlb}
Taslimipoor, S., Bahaadini, S., Kochmar, E.: {MTLB}-{STRUCT} @parseme 2020:
  Capturing unseen multiword expressions using multi-task learning and
  pre-trained masked language models. In: Proceedings of the Joint Workshop on
  Multiword Expressions and Electronic Lexicons. pp. 142--148. ACL, online (Dec
  2020), \url{https://www.aclweb.org/anthology/2020.mwe-1.19}

\bibitem{DBLP:journals/corr/abs-1809-03056}
Taslimipoor, S., Rohanian, O.: {SHOMA} at parseme shared task on automatic
  identification of vmwes: Neural multiword expression tagging with high
  generalisation. CoRR  \textbf{abs/1809.03056} (2018),
  \url{http://arxiv.org/abs/1809.03056}

\bibitem{waseem-hovy-2016-hateful}
Waseem, Z., Hovy, D.: Hateful symbols or hateful people? predictive features
  for hate speech detection on twitter. In: Proceedings of the {NAACL} Student
  Research Workshop. pp. 88--93. ACL, San Diego, California (Jun 2016).
  \doi{10.18653/v1/N16-2013}, \url{https://www.aclweb.org/anthology/N16-2013}

\end{thebibliography}

\end{document}